\documentclass[10pt,twocolumn,letterpaper]{article}

\usepackage{cvpr}              

\usepackage{indentfirst}
\usepackage[dvipsnames]{xcolor}
\usepackage{bm}
\usepackage{ifthen}

\newcommand{\ourmethod}{DreaMo}
\newcommand{\mparagraph}[1]{\vspace{0.1em}\noindent\textbf{#1}}
\newcommand{\mvec}[1]{\bm{#1}}
\newcommand{\mmat}[1]{\bm{#1}}
\newcommand{\norm}[2]{\left\lVert#1\right\rVert_{#2}}

\usepackage{xspace}

\def\signed #1{{\leavevmode\unskip\nobreak\hfil\penalty50\hskip2em
  \hbox{}\nobreak\hfil -- #1%
  \parfillskip=0pt \finalhyphendemerits=0 \endgraf}}
\newsavebox\mybox

\definecolor{ImproveGreen}{rgb}{0.2235, 0.7094, 0.2902}

\usepackage{color}
\definecolor{crimson}{rgb}{0.86, 0.08, 0.24}
\definecolor{gray}{rgb}{0.5,0.5,0.5}
\definecolor{green}{rgb}{0, 0.4, 0}
\definecolor{orange}{rgb}{1, 0.5, 0}
\definecolor{mahogany}{rgb}{0.75, 0.25, 0.0}
\definecolor{purple}{rgb}{0.6, 0, 0.6}
\definecolor{darkgreen}{rgb}{0, 0.4, 0}
\definecolor{frenchblue}{rgb}{0.0, 0.45, 0.73}
\definecolor{red}{rgb}{1,0,0}
\definecolor{yellow}{rgb}{1,1,0}
\definecolor{magenta}{rgb}{1,0,1}
\definecolor{pink}{rgb}{1,0.412,0.706}
\definecolor{ForestGreen}{RGB}{34,139,34}

\newcommand{\mycomment}[1]{}
\newcommand{\comment}[1]{}

\makeatletter
\DeclareRobustCommand\onedot{\futurelet\@let@token\@onedot}
\def\@onedot{\ifx\@let@token.\else.\null\fi\xspace}

\def\ie{\emph{i.e}\onedot}

\def\expect{\mathop{\mathbb{E}}}
\makeatother

\def\ie{i.e.,~}               

\newlength\paramargin
\newlength\figmargin
\newlength\subfigmargin
\newlength\secmargin
\newlength\subsecmargin
\newlength\tabmargin
\newlength\eqmargin

\long\def\ignorethis#1{}
\definecolor{crimson}{rgb}{0.86, 0.08, 0.24}
\definecolor{green}{rgb}{0, 0.5, 0.25}
\definecolor{purple}{rgb}{0.75, 0, 1}
\definecolor{orange}{rgb}{1, 0.5, 0.25}
\definecolor{yellow}{rgb}{1, 1, 0}
\definecolor{new_blue}{rgb}{0, 0.5, 1}

\newcommand {\tao}[1]{{\color{green}\textbf{Tao: }#1}\normalfont}
\newcommand {\hubert}[1]{{\color{new_blue}\textbf{Hubert: }#1}\normalfont}
\newcommand {\charles}[1]{{\color{orange}\textbf{Charles: }#1}\normalfont}
\newcommand {\mh}[1]{{\color{purple}\textbf{MH: }#1}\normalfont}
\newcommand {\fong}[1]{{\color{crimson}\textbf{Fong: }#1}\normalfont}

\newboolean{clean_comments}
\setboolean{clean_comments}{false}
\ifthenelse{\boolean{clean_comments}}{
    \renewcommand{\tao}[1]{}
    \renewcommand{\hubert}[1]{}
    \renewcommand{\charles}[1]{}
    \renewcommand{\mh}[1]{}
    \renewcommand{\fong}[1]{}
}{

}

\usepackage{soul}
\usepackage{multirow}
\usepackage{makecell}

\definecolor{cvprblue}{rgb}{0.21,0.49,0.74}
\usepackage[pagebackref,breaklinks,colorlinks,citecolor=cvprblue,urlcolor=cvprblue,linkcolor=cvprblue]{hyperref}

\def\paperID{2991} 
\def\confName{CVPR}
\def\confYear{2024}

\title{{\ourmethod}: Articulated 3D Reconstruction From A Single Casual Video}
\author{
Tao Tu$^{1}$\textsuperscript{*} \quad Ming-Feng Li$^{2}$\textsuperscript{*} \quad Chieh Hubert Lin$^3$ \quad Yen-Chi Cheng$^4$ \\ Min Sun$^{1,5}$ \quad Ming-Hsuan Yang$^3$
\vspace{15pt} \\
$^1$National Tsing Hua University \quad $^2$Carnegie Mellon University \\ $^3$University of California, Merced \quad $^4$University of Illinois Urbana-Champaign \quad $^5$Amazon
}

\begin{document}
\twocolumn[{%
\maketitle
\begin{center}
    \centering
    \captionsetup{type=figure}
    \includegraphics[width=\textwidth]{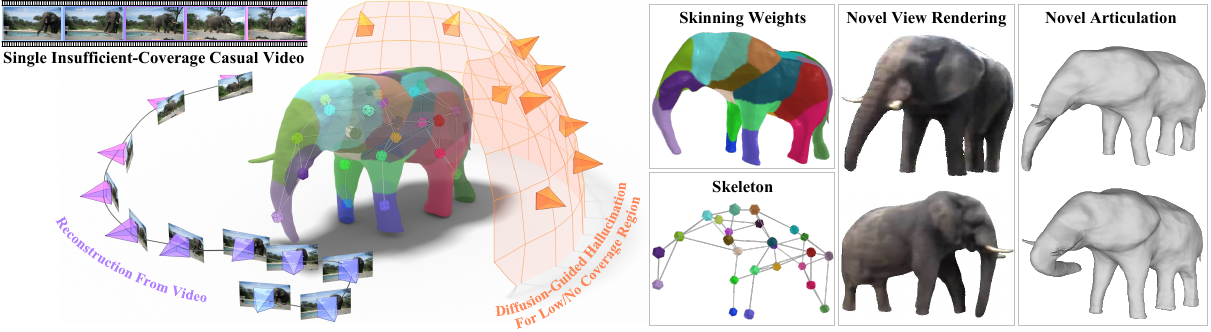}
    \vspace{-1.5em}
    \captionof{figure}{
        Casual everyday videos often lack sufficient view coverage of the subject, posing a challenge for existing articulated shape reconstruction methods.
        In contrast, {\ourmethod} not only learns deformable 3D reconstruction from visible viewpoints but also hallucinates the invisible regions with a view-conditioned diffusion model.
        As a result, {\ourmethod} achieves high-quality rendering of novel views, detailed geometry reconstruction, and provides interpretable and controllable skeletons with skinning weights.
    }
\end{center}%
}]

\setlength{\abovedisplayskip}{1.5mm}
\setlength{\belowdisplayskip}{1.5mm}
\setlength{\abovedisplayshortskip}{0pt}
\setlength{\belowdisplayshortskip}{0pt}

{\let\thefootnote\relax\footnotetext{\noindent\textsuperscript{*} indicates equal contribution.}

\begin{abstract}
Articulated 3D reconstruction has valuable applications in various domains, yet it remains costly and demands intensive work from domain experts.
Recent advancements in template-free learning methods show promising results with monocular videos.
Nevertheless, these approaches necessitate a comprehensive coverage of all viewpoints of the subject in the input video, thus limiting their applicability to casually captured videos from online sources.
In this work, we study articulated 3D shape reconstruction from a single and casually captured internet video, where the subject's view coverage is incomplete.
We propose {\ourmethod} that jointly performs shape reconstruction while solving the challenging low-coverage regions with view-conditioned diffusion prior and several tailored regularizations.
In addition, we introduce a skeleton generation strategy to create human-interpretable skeletons from the learned neural bones and skinning weights.
We conduct our study on a self-collected internet video collection characterized by incomplete view coverage.
{\ourmethod} shows promising quality in novel-view rendering, detailed articulated shape reconstruction, and skeleton generation.
Extensive qualitative and quantitative studies validate the efficacy of each proposed component, and show existing methods are unable to solve correct geometry due to the incomplete view coverage.
\vspace{-1em}
\end{abstract}
\section{Introduction}
\label{sec:intro}
Articulated 3D models have extensive applications in movie production, gaming, and virtual reality.
%
These models offer users flexible motion controls, making them highly suitable for content creation across various scenarios and specifications.
However, the manual creation of such models is expensive and time-consuming, while the quality heavily depends on artists' skill, often leading to shapes and appearances that deviate from realism.
Therefore, there are ongoing explorations into extracting articulated 3D models directly from video data due to its high accessibility on the Internet and the low hardware requirements.

Retrieving a 3D model from casual videos without constraints is a challenging and ill-conditioned problem.
The dynamic movement of the subject hampers triangulation, subsequently introducing complexity to the geometry extraction.
In addition, self-occlusion often impedes the retrieval of crucial geometry and motion cues necessary for full-shape reconstruction.
Therefore, recent methods~\cite{jiang2022neuman,zhang2021editable} often utilize parametric template models from existing 3D scans of humans and animals to regulate and guide the geometry of invisible surfaces.

While effective, collecting 3D scans for arbitrary object categories and wildlife is challenging.
BANMo~\cite{yang2022banmo} represents a recent endeavor to reconstruct the non-rigid 3D model from videos without template shape priors.
Despite BANMo demonstrating promising reconstruction quality when providing multiple videos of the same subject, we observed that it requires dense camera coverage of the same subject from multiple video sequences.
%
This requirement limits the application of retrieving 3D shapes from Internet videos or casual video captures.
The real-world videos are not dedicated to 3D shape reconstruction and, thus often have insufficient view coverage of the subject from diverse angles.
Through experiments, we show that existing 3D reconstruction methods cannot handle such types of videos.


To address the aforementioned issues, we propose {\ourmethod}, a template-free 3D articulated shape reconstruction framework tailored with joint reconstruction and hallucination.
%
%
%
{\ourmethod} simultaneously reconstructs 3D shape using neural radiance field~\cite{mildenhall2021nerf}, and hallucinates plausible geometry of invisible regions using the diffusion prior~\cite{rombach2022stablediffusion,poole2022dreamfusion,liu2023zero123}.
We analyze several design choices and show that careful parameter selection during distilling information from the diffusion model is critical for preserving high-quality surface texture (\Cref{sec:approach-view-prior}).
In addition, we introduce three regularization techniques to stabilize the learned neural bones, mitigating the generation of eccentric bumps and fragmented structures (\Cref{sec:approach-learning}).
To further enhance the interpretability and controllability of the learned 3D model, we propose a simple strategy for generating a skeleton based on the neural bones and the learned radiance field (\Cref{sec:approach-skeleton-gen}).

To validate the performance of {\ourmethod} and the limitations of existing methods on diverse animal species under different capture settings, we collect a set of short video clips with insufficient view coverage from the Internet.
Through extensive quantitative and qualitative comparisons, we show {\ourmethod} produces more plausible geometry, texture, and skeleton, compared to existing state-of-the-art approaches in articulated 3D reconstruction.
Our ablation study on each proposed design further supports the significance of these design choices.

\section{Related Work}
\label{sec:related-work}

\mparagraph{Model-based reconstruction.}
Model-based methods~\cite{badger20203dbird,biggs2020leftdogout,kocabas2020vibe,zuffi2019three,jiang2022neuman,liu2021neuralactor} build articulated 3D models from images or videos by utilizing prior parametric models~\cite{pavlakos2019expressive,xiang2019monocular,zuffi2018lions,loper2023smpl,zuffi2017smal} derived from an extensive collection of 3D scans of humans or toy animals.
Despite achieving remarkable reconstruction results, collecting 3D scan data is practically challenging, particularly for wildlife animals.

\mparagraph{Image-based reconstruction.}
Due to the ease of obtaining monocular image data from the Internet, prior studies focus on learning 3D shapes from images with weak 2D supervision, such as keypoints or silhouettes.
Some approaches~\cite{wu2023magicpony, jakab2023farm3d, goel2020ucmr, kanazawa2018cmr, kokkinos2021ttp, li2020self, ye2021shelf, kulkarni2020acsm} learn a category-specific model from an image collection and perform test-time 3D shape reconstruction using a single image.
Additionally, part-based methods~\cite{yao2022lassie, yao2023hilassie} assemble 3D parts to construct articulated 3D models from a limited Internet image collection, a scenario akin to ours.
However, unlike video data, images lack temporal information and smooth transitions across video frames, which are both valuable cues for 3D reconstruction.

\mparagraph{Video-based reconstruction.}
Video-based methods~\cite{wu2023dove, li2020vmr, jafarian2021social,bregler2000recovering,gotardo2011non,kong2019deepnrsfm,kumar2020non,kumar2017hide,yang2021viser,yang2021lasr} for articulated 3D reconstruction can leverage the temporal information inherent in given video sequences.
%
%
Inspired by the promising outcomes in novel-view synthesis research~\cite{mildenhall2021nerf, park2021nerfies,lin2021barf,pumarola2021dnerf,liu2023robust,gao2021dynamic}, recent 3D articulation reconstruction methods employ differential rendering to minimize reconstruction loss.
While some of these methods can generate plausible 3D models, they require specific inputs such as multi-view videos~\cite{peng2021animatablenerf, zhang2021editable}, predefined 3D skeletons~\cite{weng2022humannerf, weng2020vid2actor, peng2021animatablenerf, su2021anerf}, or 3D rest-pose point clouds~\cite{su2023npc}.
Among them, BANMo~\cite{yang2022banmo} shows promising 3D reconstruction results for articulated objects solely using video data.
However, it still demands several videos featuring dense camera coverage of the same subject, limiting its feasibility for a single casual video.

\mparagraph{Distillation from diffusion models.}
2D diffusion models~\cite{rombach2022stablediffusion,saharia2022photorealistic,zhang2023controldiffusion,ramesh2021dalle} have demonstrated promising results in generating realistic 2D images.
Building upon this foundation, recent works~\cite{poole2022dreamfusion,lin2023magic3d,lazova2019360,metzer2023latentnerf,raj2023dreambooth3d,richardson2023texture,singer2023text} achieve 3D model reconstruction by employing a pretrained 2D diffusion model as a prior for view synthesis~\cite{poole2022dreamfusion, wang2023scorejacobian}.
To gain control of the camera viewpoint over the 2D diffusion model, Zero-1-to-3~\cite{liu2023zero123} finetunes the diffusion model on a synthetic dataset~\cite{deitke2023objaverse} and demonstrates the zero-shot ability to generate novel views for a specified subject.
Such view control capability allows us to harness the generative power of the 2D diffusion model by imagining parts of the subject that are not observed in the input video, enabling us to improve the reconstruction of the target subject.

\begin{figure*}[t]
\begin{center}
\includegraphics[width=\textwidth]{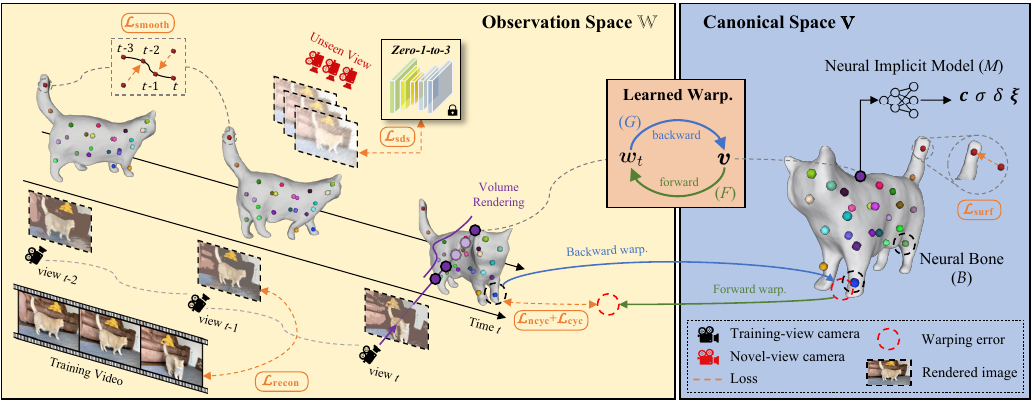}
\end{center}
\vspace{-1.5em}
\caption{
  \textbf{\ourmethod} is a template-free articulated 3D shape reconstruction framework, which jointly performs training-view reconstruction and unseen-region hallucination.
  It learns a rest-pose neural 3D model ($M$) using a neural implicit function in the canonical space ($\mathbb{V}$).
  A forward warping model ($F$) transforms this 3D model into observation space ($\mathbb{W}$), where the video frames supervise the model to capture time-dependent motions.
  %
  Conversely, a backward warping model ($G$) performs the inverse operation of $F$.
  {\ourmethod} uses Zero-1-to-3 to hallucinate and complete the unseen regions.
  Meanwhile, we introduce several regularization terms ($\mathcal{L}_{\text{ncyc}}$, $\mathcal{L}_{\text{surf}}$, $\mathcal{L}_{\text{smooth}}$) to improve the placement of neural bones and reduce geometric artifacts.
}
\vspace{-1em}
\label{fig:method}
\end{figure*}

\section{Approach}
\label{sec:approach}
We tackle the problem of articulated 3D shape reconstruction from a single video. 
The recovered 3D shape encapsulates the time-dependent motions observed in the video, and the learned neural bones with skinning weights support novel articulations controlled by the users.
%
Similar to BANMo~\cite{yang2022banmo}, our {\ourmethod} learns a rest-pose neural 3D model represented by a neural implicit function in the time-invariant canonical space $\mathbb{V}$ and a separate time-dependent warping implicit function deforms the canonical space features to the observation space $\mathbb{W}$.
All the image supervision involving volume rendering~\cite{mildenhall2021nerf} occurs in the observation space. 
%
\Cref{fig:method} shows the overview of {\ourmethod}. 


\subsection{Articulated 3D Reconstruction Model}
\label{sec:approach-3d-model}
\mparagraph{Neural implicit model in canonical space ($M$).}
The neural implicit function $M$ in the canonical space represents a rest pose of the 3D reconstruction target.
Following previous works in neural rendering~\cite{mildenhall2021nerf,yariv2021volsdf,yang2022banmo}, we model color $\mvec{c}$, signed distance function (SDF) $\delta$, density $\sigma$, and semantic feature $\mvec{\xi}$ by neural implicit functions with
\begin{equation}
\mvec{c}, \sigma, \delta, \mvec{\xi} = M(\mvec{v}),
\label{eq:canonical-model}
\end{equation}
where $\mvec{v} \in \mathbb{V}$ is a coordinate in canonical space.
\comment{
\begin{equation}
\mvec{c}, \sigma, \delta, \mvec{\xi} = M(\mvec{v}, \mvec{d}),
\label{eq:canonical-model}
\end{equation}
where $\mvec{v} \in \mathbb{V}$ is a 3D coordinate in canonical space and $\mvec{d}$ is the view direction of a radiance while performing volume rendering.
}
We implement $M$ with Multi-Layer Perceptron (MLP) and follow VolSDF~\cite{yariv2021volsdf} to convert the SDF values $\delta$ to density $\sigma$ by the cumulative distribution function of Laplacian distribution.

\mparagraph{Linear blend skinning.}
Similar to BANMo~\cite{yang2022banmo}, in order to provide explicit representation for users to articulate the reconstructed 3D model, we use Gaussian ellipsoids to represent a set of neural bones $\mathcal{B}=\{\mvec{\mu}_b, \mmat{\Sigma}_b\}_{b=1}^{B}$, where $B$ is the number of neural bones, $\mvec{\mu}_b$ is the bone position, and $\mmat{\Sigma}_b$ represents the bone orientation and scale.
The transformation of a skin point at coordinate $\mvec{v}$ is determined by weighting over the $B$ bone transformations based on the skinning weights, following the linear blend skinning method~\cite{skinningcourse2014}.
A higher skinning weight implies that the transformation of the associated bone will introduce larger deformation to the skin point $\mvec{v}$.
Specifically, we determine the per-bone skinning weights for $\mvec{v}$ by the Mahalanobis distance between $\mvec{v}$ and the $b$-th neural bone, which are subsequently normalized using a softmax layer.

\mparagraph{Time-dependent warping models ($F$ and $G$).}
To represent the time-dependent motion of the reconstructed subject, we use a forward warping model $F$ to transform points in canonical space to the observation space of time $t$ with $\mvec{w} = F(\mvec{v}, t) \in \mathbb{W}$.
We also jointly learn a backward warping function $\mvec{v} = G(\mvec{w}, t)$ to approximate the inverse of $F$.
More specifically, we implement the warping functions as a combination of linear blend skinning that blends a set of bone transformations $\{\mmat{J}_{t,b}\}_{b=1}^{B}$ for modeling the subject's deformation, and a global transformation $\mmat{P}_t(\cdot) \in \text{SE(3)}$ that represents the camera transformation.

\mparagraph{Volume rendering.}
We use volume rendering~\cite{yang2022banmo, mildenhall2021nerf} to render images at different time steps $t$.
To render a pixel $u$ at time $t$, we cast a ray from a posed pinhole camera at time $t$, and sample a sequence of 3D points $\mvec{w}_{i}^{u}$ along the ray in the observation space, where $i$ is the index of the point on the ray.
We first obtain the point-wise canonical space color $\mvec{c}_{i}^{u,t}$, density $\sigma_{i}^{u,t}$, SDF value $\delta_{i}^{u,t}$, and semantic features $\mvec{\xi}_{i}^{u,t}$ in \Cref{eq:canonical-model} with
\begin{equation}
    \mvec{c}_{i}^{u,t}, \sigma_{i}^{u,t}, \delta_{i}^{u,t}, \mvec{\xi}_{i}^{u,t} = M\left( \, G(\mvec{w}_{i}^{u}, t)\right) \, .
\end{equation}
\comment{
\begin{equation}
    \mvec{c}_{i}^{u,t}, \sigma_{i}^{u,t}, \delta_{i}^{u,t}, \mvec{\xi}_{i}^{u,t} = M\left( \, G(\mvec{w}_{i}^{u}, t), \,\, \mvec{d}^{u} \right) \, .
\end{equation}
}
Subsequently, the pixel color $\mvec{c}^{u,t}$ and silhouette $o^{u,t}$ in the camera space are obtained using volume rendering:
\begin{equation}
\label{eq:vol-render}
    \mvec{c}^{u,t} = \sum_i\tau_i\mvec{c}_{i}^{u,t}, \,\,\,\,\,\, o^{u,t} = \sum_i\tau_i,
\end{equation}
where $\tau_i=\alpha_i{\prod}_{j=1}^{i-1}(1-\alpha_j)$ are the weights along the ray, $\alpha_i=1-\exp(-\sigma_{i}^{u,t}\Delta_{i})$ are the alpha compositing values, and $\Delta_{i}$ is the distance between the $i$-th sample and the subsequent sample along the ray.
Likewise, the camera space semantic features $\mvec{\xi}^{u,t}$ are rendered by replacing the pixel color $\mvec{c}_{i}^{u,t}$ 
with semantic feature $\mvec{\xi}_{i}^{u,t}$ 
in \Cref{eq:vol-render}.

\paragraph{3D model manipulation.}
%
To articulate the rest-pose 3D model reconstructed by {\ourmethod} into a user-defined pose, the user only needs to specify the bone transformations from the rest to the target pose.
%
%
With these bone transformations, for any points in the space, we can derive the skinning weights contributed by each bone and determine the deformation of all the points on the object's surface.
%
In practice, we convert the implicit model in canonical space into an explicit mesh representation $\mathcal{V} = \{\mvec{\nu}_i \in \mathbb{R}^3\}$ using the marching cube algorithm.
Subsequently, we morph the mesh by deforming each vertex $\mvec{\nu}_i$ to the target pose.
Finally, the color of each vertex is determined by querying the implicit model in canonical space.

%
%

\subsection{View-Conditioned Diffusion Model as Prior}
\label{sec:approach-view-prior}
One of our main challenges is recovering appropriate geometry for unseen surfaces of the deformable subject.
In order to support the low-coverage or unseen view angles, we use Zero-1-to-3~\cite{liu2023zero123} to synthesize the novel view conditioning on a source image of the same time step and the camera pose of the novel view.
These synthetic supervisions are distilled into the articulated 3D reconstruction model using Score Distillation Sampling (SDS)~\cite{poole2022dreamfusion}.

However, we found naively updating all the trainable parameters with SDS will destroy the high-frequency details of the reconstructed surface texture (\Cref{fig:ablation-params}).
This leads to a blurry and over-saturated appearance similar to the relevant works in 3D asset generation~\cite{poole2022dreamfusion,liu2023zero123,wang2023scorejacobian}.
We hypothesize this is due to the randomness of the diffusion model, which causes the novel views inconsistent with the training views and makes the radiance field unable to converge.

As a remedy, recall that we only intend to enhance the geometry of the low-coverage surface, we discover making the SDS gradients only update the geometry-relevant parameters can achieve the goal without compensating the surface texture.
In practice, we only update the parameters of the neural bones $\mathcal{B}$.
The ablation study in \Cref{tab:ablation-metrics} and \Cref{fig:ablation-params} shows the effectiveness of this parameter selection.

\subsection{Learning}
\label{sec:approach-learning}
Our end-to-end full training objective is a weighted sum of reconstruction loss $\mathcal{L}_{\text{recon}}$, SDS loss $\mathcal{L}_{\text{sds}}$, and regularization terms $\mathcal{L}_{\text{cyc}}$, $\mathcal{L}_{\text{ncyc}}$, $\mathcal{L}_{\text{surf}}$, $\mathcal{L}_{\text{smooth}}$.

\mparagraph{Reconstruction Loss.}
The reconstruction loss minimizes the discrepancies between {\ourmethod}'s rendered images and the ground-truth images in training views.
We also train the model to predict optical flow as it aids in learning correct deformation by providing approximated pixel correspondence across time.
For a pixel $u$ at time $t$, the optical flow $\mvec{f}_{t \to t'}^{u}$ represents the warping vector between $u$ and its new pixel location $u'$ at $t'$ after object deformation and viewpoint transformation.
Such $ u'$ is computed by backward warping the observation space points at $t$ to the canonical space, which is time-invariant, forward warping to the observation space at $t'$, and then applying perspective camera projection $\mathrm{\Gamma}_{t'}(\cdot)$ derived from the learned global transformation $\mmat{P}_{t'}$ and the camera intrinsic at time $t'$.
The resulting $u'$ can be expressed as
$
    u' = \sum_{i}\tau_i\mathrm{\Gamma}_{t'}\left(F\left(G(\mvec{w}_{i}^{u}, t)\right), t'\right).
$
Finally, the reconstruction loss is
\begin{equation}
\begin{split}
\label{eq:reconstruction-loss}
\mathcal{L}_{\text{recon}} = \expect_{(u, t, t')}
& \| \hat{\mvec{c}}^{u, t} - \mvec{c}^{u, t}\|_2 +
\| \hat{\mvec{\xi}}^{u, t} - \mvec{\xi}^{u, t}\|_2 + \\[-2mm]
& \| \hat{o}^{u, t} - o^{u, t} \|_2 +
\| \hat{\mvec{f}}_{t \to t'}^{u} - \mvec{f}_{t \to t'}^{u} \|_2 \,\, ,
\end{split}
\end{equation}
where we randomly sample $N$ combinations of $(u, t, t')$ from all unique combinations for each training iteration.

\mparagraph{Regularization.}
To encourage the inverse relationship between the forward and backward warping functions, we employ a training-view cyclic consistency loss $\mathcal{L}_{\text{cyc}}$ similar to BANMo.
In addition, since our problem involves large regions unseen by the training views, we introduce the novel-view cyclic consistency loss $\mathcal{L}_{\text{ncyc}}$:
\begin{align}
\label{eq:reg-warp-consistency}
\mathcal{L}_{\text{cyc}} &= \sum_{m}\tau_{m}\norm{\mvec{w}_{m} - F(G(\mvec{w}_{m}, t), t)}{2}, \\
\mathcal{L}_{\text{ncyc}} &=  \sum_{n}\tau_{n}\norm{\mvec{w}_{n} - F(G(\mvec{w}_{n}, t), t)}{2},
\end{align}
where \(\mvec{w}_{m}\) and \(\mvec{w}_{n}\) are coordinates of points on the training-view and novel-view rays, respectively.

In addition, we found learning the neural bones without additional constraints allows them to scatter all over the space. 
This creates unusual geometry and floaters in the regions with insufficient view coverage.
Therefore, we introduce a surface constraint loss 
$\mathcal{L}_{\text{surf}}$ to keep the bones beneath the SDF surface:
\begin{equation}
\label{eq:reg-within-surface}
\mathcal{L}_{\text{surf}} = \| \, \text{max} \{ \delta, 0\} \, \|_2
\end{equation}
where $\delta$ is the SDF value of the bones.
Meanwhile, we observe that the learned transitions of bones in the low-coverage or self-occluded regions often exhibit unnatural jiggles.
These unnatural motions also frequently produce broken geometry and floaters.
For this, we design a smooth transition loss $\mathcal{L}_{\text{smooth}}$ to encourage both the translation and rotation of each bone to have steady change over time:
\begin{equation}
\label{eq:reg-smooth-transition}
\begin{split}
\mathcal{L}_{\text{smooth}} = \sum^{B,T-1}_{b=1,t=1} \frac{\mathrm{ang}(\mmat{R}^{t}_{b}, \mmat{R}^{t+1}_{b})
+ \norm{\mvec{s}^{t}_{b} - \mvec{s}^{t+1}_{b}}{2}}{{B(T-1)}},
\end{split}
\end{equation}
where \(\left(\mmat{R}^{t}_{b}|\mvec{s}^{t}_{b}\right)=\mmat{J}^{t}_{b}\), and 
we compute the relative angle of rotations with
\(\mathrm{ang}(\mmat{R}_{1}, \mmat{R}_{2})=\arccos((\mathrm{tr}(\mmat{R}_{1}\mmat{R}_{2}^{\intercal})-1)/2)\).

\subsection{Skeleton Generation}
\label{sec:approach-skeleton-gen}

\begin{figure}[t]
\begin{center}
\includegraphics[width=\linewidth]{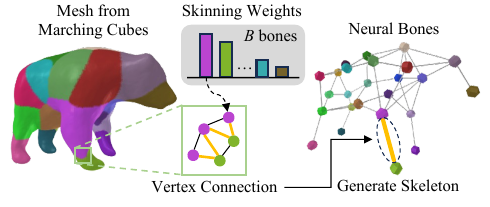}
\end{center}
\vspace{-2em}
\caption{
  \textbf{Skeleton generation strategy.}
  Initially, we extract the rest-pose surface of the neural implicit 3D model using marching cubes.
  Subsequently, each vertex is assigned to a neural bone based on the maximum skinning weights.
  An edge between bones is established if there exists a sufficient vertex connection.
}
\vspace{-1em}
\label{fig:skeleton-generation}
\end{figure}

In this section, we describe how {\ourmethod} generates a skeleton from the learned neural bones, and illustrate the strategy in \Cref{fig:skeleton-generation}.
The skeleton defines the pairwise relationship between bones, indicating whether two bones have close interactions to jointly control a shared set of points on the SDF surface or the vertices of the mesh converted from the SDF.
%
We assess such a relationship using the skinning weights, which define the contribution weights of the neural bones to each particular point in the space.

With the canonical space model $M$ learned with the objectives described in \Cref{sec:approach-learning}, we can extract the rest-pose mesh by running marching cubes. 
We denote the mesh as a collection of vertices $\mathcal{V} = \{\mvec{\nu}_i \in \mathbb{R}^3\}$ and faces $\Lambda = \{\lambda_j\}$ where each element $\lambda_j$ denotes a triangle face formed by three vertices in $\mathcal{V}$.
Additionally, we extract the skinning weight of each vertex based on its coordinates and then identify the bone $b\in\mathcal{B}$ with the highest skinning weight to control that vertex.
After iterating through all the vertices, we obtain clusters of vertices, each assigned with the corresponding bone ID.
Our next target is to find the bone pairs that control a significant number of shared faces, which implies the two bones will deform a considerable number of common faces during articulation.
We traverse all the face edges of the mesh.
For each pair of bones $(b_1, b_2)$, we count the number of edges that connect two vertices assigned to these two bones.
Finally, we decide whether each pair of bones is connected using such surface connectivity count and simple threshold to remove extreme outliers (e.g., two bones sharing only a single face).

\section{Experimental Results}
\label{sec:experiment}
\mparagraph{Implementation details of {\ourmethod}.}
Our canonical space implicit model, warping models, neural bones, and global transformation network are all MLPs with residual connections similar to NeRF~\cite{mildenhall2021nerf}.
To model the fine-grain deformations, we learn another MLP to predict delta skinning weights and add it to the linear skinning weights before the softmax layer as in BANMo~\cite{yang2022banmo}.
To obtain the SDS loss, we randomly sample a relative azimuth between $[-90^{\circ}, 90^{\circ}]$ and a relative elevation in $[-10^{\circ}, 45^{\circ}]$ as the viewpoint condition of Zero-1-to-3~\cite{liu2023zero123}.
%
%
%
During volume rendering, we sample 64 points along the radiance for most of the losses, except the SDS loss only uses 16 samples to reduce memory consumption.
Throughout the experiments, the weights of our loss terms $\mathcal{L}_{\text{recon}}$, $\mathcal{L}_{\text{sds}}$, $\mathcal{L}_{\text{cyc}}$, $\mathcal{L}_{\text{ncyc}}$, $\mathcal{L}_{\text{surf}}$, $\mathcal{L}_{\text{smooth}}$ are $10^{-1}$, $10^{-4}$, $10^{-2}$, $1$, $10^{-1}$, and $10^{-2}$, respectively.
We reduce the frequency of computationally expensive losses to accelerate the training, specifically, we update $\mathcal{L}_{\text{ncyc}}$ every 3 iterations and $\mathcal{L}_{\text{sds}}$ every 10 iterations.
%

\begin{figure*}[t]
\begin{center}
\includegraphics[width=\textwidth]{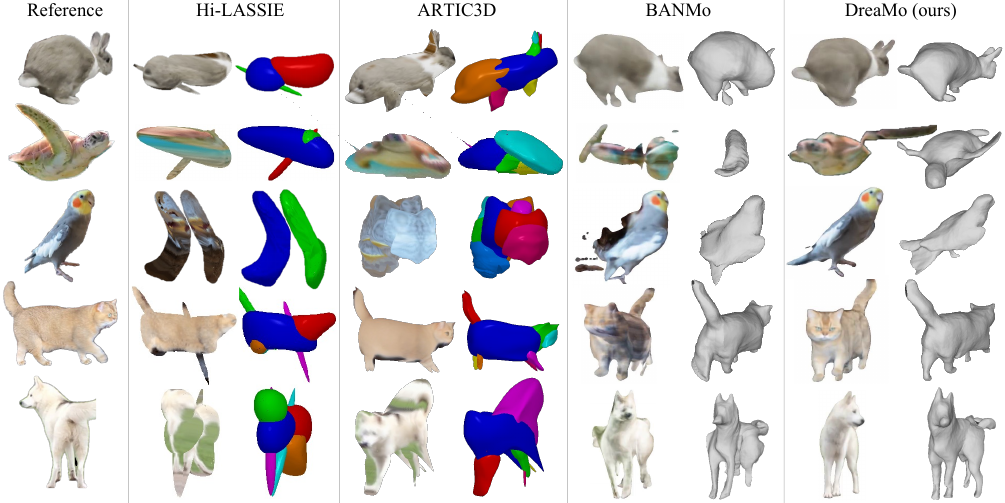}
\end{center}
\vspace{-1.5em}
\caption{
  \textbf{3D reconstruction comparison among state-of-the-arts.}
  {\ourmethod} generates more complete 3D shapes and maintains richer high-frequency details in the novel view rendering.
  %
  For each method, we show the rendered novel-view image and the reconstructed shape.
  %
  \vspace{-2em}
}
\label{fig:recon-base}
\end{figure*}

\mparagraph{Datasets.}
%
We collected 42 animal video clips with diverse species and insufficient view coverage from the Internet.
To ensure the videos satisfy our challenging insufficient-coverage scenario, we compute and ensure the videos have low azimuth viewpoint coverage during the data collection.
We compute the azimuth coverage by evenly dividing the azimuth of canonical space into 36 equally angle bins, finding the bins covered by any learned global transformation $\mmat{P}_t$ (see Section~\ref{sec:approach-3d-model}), then determining the occupancy ratio.
The average azimuth viewpoint coverage is 31\%.
%
%
%
%
Our video set contains 28 different animal species, and each video only contains one animal subject.
%
The average duration of the videos is $15.7$ seconds.
We follow the data preprocessing protocol from our baseline BANMo.
In particular, an intermediate step utilizes optical flow~\cite{yang2021vcnplus} to filter out frames with small motions.
The final averaged number of video frames utilized by our {\ourmethod} and BANMo is $124.4$.
%

\mparagraph{Metrics.}
We consider two aspects: (a) the consistency between the input video and the reconstructed 3D model, and (b) the visual quality of the re-rendered images.

For the consistency evaluation, we only consider semantic consistency. 
In our problem setting, the pixel-level reconstruction metrics in novel views are not applicable because the majority of view angles are inherently unobserved within the input video.
Furthermore, it is infeasible to hold out testing video frames from training, 
as the subject's articulations in test frames are unknown to {\ourmethod} and non-trivial to discover.
It is difficult to find the correct pose alignment between the canonical space model with the unseen novel views.
%
%
To measure the semantic consistency, we use the image encoder of CLIP~\cite{radford2021clip} to extract image features and compare the cosine similarity between the training and the rendered video frames.
Specifically, we compute the cosine similarity in two different strategies: \textit{exhaustive} that compares each rendered novel view with all the training video frames, and \textit{per-time} that only compares the video frames at the same time step.
%
%
%
%

For quality assessment, we use Kernel Inception Distance (KID)~\cite{binkowski2018mmdgans, obukhov2020torchfidelity}, which is a popular quality measurement metric in generative modeling research specialized for small sample sizes.
For each time step, we render a novel view at the same time step, resulting in an image set at the same size as the number of training video frames.
We calculate the KID between these two sets of images, and then report both the mean and the standard deviation.


\begin{figure}[!ht]
\begin{center}
\includegraphics[width=\linewidth]{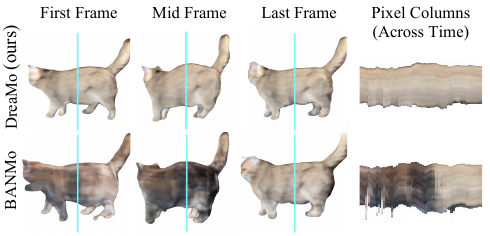}
\end{center}
\vspace{-1em}
\caption{
    \textbf{DreaMo maintains better visual consistency.}
    We render the video from the same novel-view angle, retrieve a pixel column from each time step ({\color{cyan} cyan} line), and aggregate them on the right. 
    BANMo often shows a drastic color shift over time.
}
\label{fig:consistency}
\end{figure}

\begin{table}[!ht]
\small
\centering
\caption{
    \textbf{{\ourmethod} maintains better semantical consistency and better visual quality.}
    We measure the semantical consistency with CLIP similarity and quantify visual quality using KID. $\dag$ denotes evaluated on a subset of 11 videos.
}
\vspace{-0.5em}
    \begin{tabular}{@{}lccc@{}}
    \toprule
    \multirowcell{2}[-0.6ex][l]{Method} & \multicolumn{2}{c}{CLIP ($\uparrow$)} & \multicolumn{1}{c}{KID ($\downarrow$)} \\
    \cmidrule(l){2-3} \cmidrule(l){4-4}
    & \multicolumn{1}{c}{Exhaustive} & \multicolumn{1}{c}{Per-time} & \multicolumn{1}{c}{Mean $\pm$ Stddev} \\
    \midrule
    Hi-LASSIE~\cite{yao2023hilassie} & \multicolumn{1}{c}{0.742} & \multicolumn{1}{c}{0.744} & \multicolumn{1}{c}{0.0876 $\pm$ 0.0021} \\
    BANMo~\cite{yang2022banmo} & \multicolumn{1}{c}{0.792} & \multicolumn{1}{c}{0.797} & \multicolumn{1}{c}{0.0576 $\pm$ 0.0021} \\
    {\ourmethod} (ours) & \multicolumn{1}{c}{\textbf{0.813}} & \multicolumn{1}{c}{\textbf{0.817}} & \multicolumn{1}{c}{\textbf{0.0488} $\pm$ 0.0020} \\
    \midrule
    ARTIC3D\textsuperscript{\dag}~\cite{yao2023artic3d} & \multicolumn{1}{c}{0.774} & \multicolumn{1}{c}{0.778} & \multicolumn{1}{c}{0.0822 $\pm$ 0.0036} \\
    {\ourmethod}\textsuperscript{\dag} (ours) & \multicolumn{1}{c}{\textbf{0.866}} & \multicolumn{1}{c}{\textbf{0.870}} & \multicolumn{1}{c}{\textbf{0.0350} $\pm$ 0.0024} \\
    \bottomrule
    \end{tabular}
    \vspace{-1em}
\label{tab:reconst-metrics}
\end{table}

\subsection{3D Reconstruction}
\label{sec:experiment-recon}
In \Cref{fig:recon-base}, we compare with state-of-the-art methods in articulated 3D shape reconstruction.
 
\mparagraph{Video-based methods.} 
Our most relevant baseline is the state-of-the-art video-based 3D reconstruction method, BANMo~\cite{yang2022banmo}, which has shown outstanding quality when input videos have a very dense camera coverage.
We use the latest implementation~\cite{gh-lab4d} from the author and ensure the fairness of comparisons by adding back the missing sinkhorn loss mentioned in the original paper of BANMo~\cite{yang2022banmo}.
In \Cref{fig:recon-base}, BANMo shows blurry reconstruction with fragmented geometry due to insufficient view coverage in certain regions.
With the same setting, {\ourmethod} successfully maintains clean and plausible geometry in the novel-view angles.
%
In \Cref{fig:consistency}, BANMo struggles to maintain appearance consistency across time in these low-coverage regions.
In \Cref{fig:skeleton}, we show our regularizations result in better neural bone placements, which prevents the model from creating irregular shape artifacts.
The quantitative comparisons in \Cref{tab:reconst-metrics} also indicate our {\ourmethod} maintains better semantic consistency from all angles while presenting better visual quality in KID measurement.
Please refer to the supplementary material for more reconstruction results.

\begin{figure*}[t]
\begin{center}
\includegraphics[width=0.98\textwidth]{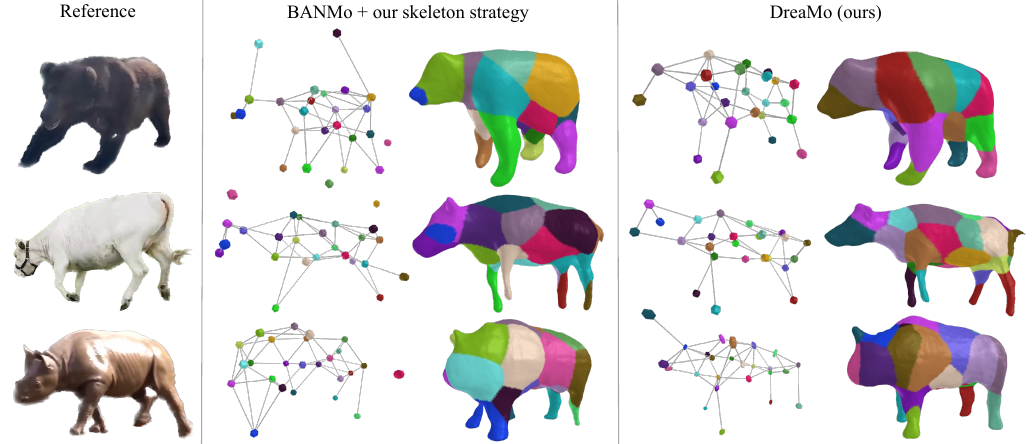}
\end{center}
\vspace{-1.5em}
\caption{
  \textbf{{\ourmethod} generates intuitive skeletons aligned with mesh.}
  We apply our skeleton generation strategy to both BANMo and {\ourmethod}, then show the skeleton and the mesh converted from the canonical model.
  The face colors of the mesh denote the bone ID with the highest skinning weight.
  Our {\ourmethod} better aligns with the 3D shape without leftover bones drifting far away from the object surface.
  \vspace{-1em}
}
\label{fig:skeleton}
\end{figure*}

\mparagraph{Image-based methods.} 
Besides BANMo, we further compare our {\ourmethod} with LASSIE~\cite{yao2023hilassie} and ARTIC3D~\cite{yao2023artic3d}, which are weakly relevant works in template-based articulated 3D-shape reconstruction from image collection of the same animal species.
These methods consider fitting the appearance of an articulated template shape with images of the same animal species with diverse viewing angles.
Despite not having the same level of visual diversity as an image collection, we put our best effort into examining these methods using video frames in replace of the image collection.
%
%
%
We run Hi-LASSIE using the official codebase~\cite{gh-hilassie}.
To satisfy the additional requirements of the skeleton initialization from Hi-LASSIE, for each video, we manually select a frame where most of the subject's parts are visible and suitable for skeleton initialization.
%
As of ARTIC3D, we sought the authors to run their algorithm on an 11-video subset from our dataset, considering the computational costs.

In \Cref{fig:recon-base}, both Hi-LASSIE and ARTIC3D often fail to solve the correct correspondence between the neural body parts and the subject in the video.
Such a problem is reinforced in their part-based optimization when each part struggles to match the appearance of the reference, thus resulting in an unrealistic appearance as a whole in the end.
In contrast, our {\ourmethod} models the entire subject using a single neural implicit field, facilitates the model to solve all the body parts at once, and results in a more consistent shape.
Moreover, {\ourmethod} leverages optical flow across video frames to better understand the pixel correspondence from the limited data, while image-based methods do not consider such information, and integrating optical flow information into their framework is non-trivial.
We also quantitatively compare with Hi-LASSIE and ARTIC3D in \Cref{tab:reconst-metrics}, our method consistently shows better semantic consistency in CLIP score and visual quality in KID.

\subsection{Skeleton Generation}
\label{sec:experiment-skeleton-gen}
We evaluate the skeletons generated by {\ourmethod} against those produced by BANMo~\cite{yang2022banmo} in \Cref{fig:skeleton}.
While BANMo~\cite{yang2022banmo} can produce reasonable object surfaces, its bone placement does not well represent the object shape, with some bones distributed far outside the object surfaces.
In contrast, since {\ourmethod} generates more accurate shapes and encourages neural bones to stay within the reconstructed surface, it results in detailed and interpretable skeletons capturing features such as limbs and heads.

\subsection{Articulating 3D Model}
\label{sec:experiment-articulate}
In \Cref{fig:articulate}, we demonstrate the controllability of {\ourmethod} by articulating the learned neural bones.
%
Following the 3D model manipulation described in~\Cref{sec:approach-3d-model}, we manually adjust the positions of the neural bones and transform the skin points (\ie vertices of the mesh) to produce novel poses.
With the accurately learned bone placement, skinning weights, and 3D shapes from {\ourmethod}, the skin points can be reasonably transitioned corresponding to the movement of the neural bones, leading to plausible results in novel poses.

\begin{figure*}[t]
\begin{center}
\includegraphics[width=.98\textwidth]{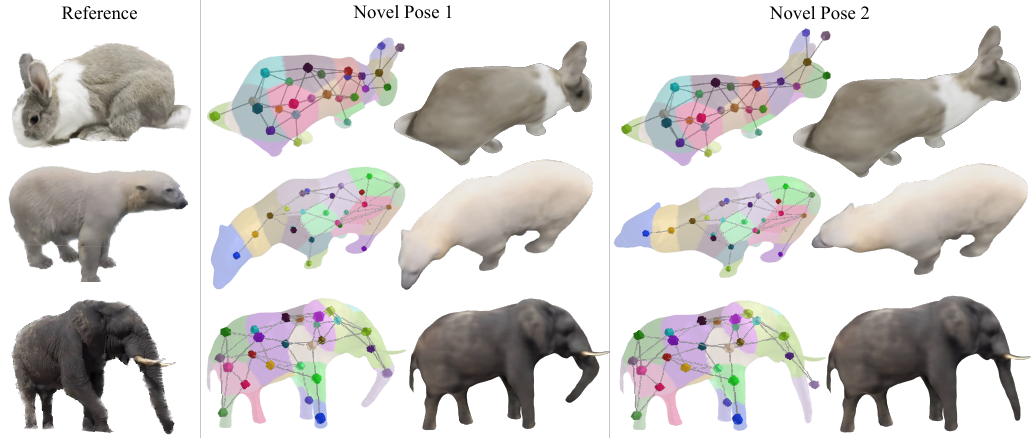}
\end{center}
\vspace{-1.5em}
\caption{
  \textbf{Controlling {\ourmethod} by manipulating generated skeletons.}
  We generate plausible 3D shapes of target objects in novel poses by controlling the generated skeletons. 
  We manually adjust the bone positions and warp the corresponding skin, i.e., the mesh vertices in the figure, to demonstrate new poses of the articulated 3D objects.
}
\label{fig:articulate}
\end{figure*}

\subsection{Ablation Study}
\label{sec:experiment-ablation}
We quantitatively analyze the contribution from different components of {\ourmethod} in \Cref{tab:ablation-metrics}.
The results show that each component contributes to the performance improvements in semantic consistency in the CLIP score and visual quality in the KID metric.
%
Among all components, the diffusion prior makes the most significant improvement by guiding {\ourmethod} in updating articulation parameters to minimize $\mathcal{L}_\text{sds}$.
Additionally, in \Cref{fig:ablation-params}, we show that updating all parameters with $\mathcal{L}_\text{sds}$ results in overly-smoothed textures.
%
This effect may potentially stem from the randomness of the denoising diffusion process, which generates view-inconsistent results that hinder the model from optimizing detailed textures.

The proposed regularization schemes also consistently improve the performance of {\ourmethod}.
These schemes help avoid placing the neural bones in empty space far from the object's surface, or sudden discontinuous transitions between frames.
These improvements also contribute to the interpretable skeletons, as demonstrated in the \Cref{sec:experiment-skeleton-gen}.


\begin{table}[t]
\small
\centering
\caption{
  \textbf{Ablation study.}
  We show the every proposed component contributes to the final improvements in semantic consistency (CLIP) and visual quality (KID).
}
\vspace{-0.5em}
\setlength{\tabcolsep}{4.5pt}
\renewcommand{\arraystretch}{0.95}
\begin{tabular}{@{}lccc@{}}
\toprule
\multirowcell{2}[-0.6ex][l]{Method} & \multicolumn{2}{c}{CLIP ($\uparrow$)} & \multicolumn{1}{c}{KID ($\downarrow$)} \\ \cmidrule(l){2-3} \cmidrule(l){4-4} 
& \multicolumn{1}{c}{Exhaust.} & \multicolumn{1}{c}{Per-time} & \multicolumn{1}{c}{Mean $\pm$ Stddev} \\ \midrule
No $\mathcal{L}_\text{ncyc}$, $\mathcal{L}_\text{smooth}$, $\mathcal{L}_\text{surf}$ & \multicolumn{1}{c}{0.799}          & \multicolumn{1}{c}{0.801} & \multicolumn{1}{c}{0.0542 $\pm$ 0.0016} \\
No $\mathcal{L}_\text{smooth}$, $\mathcal{L}_\text{surf}$                            & \multicolumn{1}{c}{0.805}          & \multicolumn{1}{c}{0.808} & \multicolumn{1}{c}{0.0507 $\pm$ 0.0019} \\
No $\mathcal{L}_\text{ncyc}$ & \multicolumn{1}{c}{0.808}          & \multicolumn{1}{c}{0.812} & \multicolumn{1}{c}{0.0535 $\pm$ 0.0021} \\
No $\mathcal{L}_\text{smooth}$ & \multicolumn{1}{c}{0.811}          & \multicolumn{1}{c}{0.815} & \multicolumn{1}{c}{0.0512 $\pm$ 0.0019} \\
No $\mathcal{L}_\text{surf}$ & \multicolumn{1}{c}{0.807}          & \multicolumn{1}{c}{0.810} & \multicolumn{1}{c}{0.0501 $\pm$ 0.0020} \\ 
No $\mathcal{L}_\text{sds}$ & \multicolumn{1}{c}{0.797}          & \multicolumn{1}{c}{0.800} & \multicolumn{1}{c}{0.0539 $\pm$ 0.0020} \\
$\mathcal{L}_\text{sds}$ for all params                                                                   & \multicolumn{1}{c}{0.794}          & \multicolumn{1}{c}{0.796} & \multicolumn{1}{c}{0.0585 $\pm$ 0.0022} \\ \midrule
{\ourmethod} (ours)                                                                  & \multicolumn{1}{c}{\textbf{0.813}} & \multicolumn{1}{c}{\textbf{0.817}} & \multicolumn{1}{c}{\textbf{0.0488} $\pm$ 0.0020} \\ \bottomrule
\end{tabular}
\vspace{-1em}
\label{tab:ablation-metrics}
\end{table}
\begin{figure}[t]
    \centering
    \begin{subfigure}[b]{0.51\linewidth}
         \includegraphics[width=\linewidth]{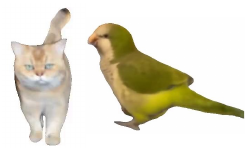}
         \caption{DreaMo (ours)}
    \end{subfigure}
    \hfill
    \begin{subfigure}[b]{0.47\linewidth}
         \includegraphics[width=\linewidth]{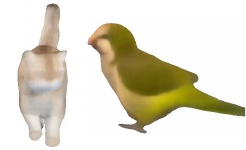}
         \caption{$\mathcal{L}_\text{sds}$ for all params}
    \end{subfigure}
    \vspace{-0.5em}
    \caption{Naively updating all the trainable parameters with SDS hinders the high-frequency details of the reconstructed texture.}
    \label{fig:ablation-params}
    \vspace{-1.2em}
\end{figure}

\section{Conclusion And Limitations}
\label{sec:conclusion}
We present {\ourmethod}, a template-free framework to reconstruct plausible articulated 3D models from a single casual video with incomplete view coverage.
%
To overcome the insufficient supervision in the unseen or low-coverage regions,
{\ourmethod} leverages the view-conditioned diffusion model to hallucinate and complete the 3D shape with a plausible and coherent appearance. 
%
Besides, we show the effectiveness of our proposed regularization schemes that improve the placement of the neural bones and reduce the irregular reconstruction artifacts.
%
We further present a simple skeleton generation strategy to transform the learned neural bones and skinning weights into interpretable skeletons.
Through extensive qualitative and quantitative experiments, we show {\ourmethod} achieves state-of-the-art quality in articulated 3D shape reconstruction in our single video setting.

Despite {\ourmethod} achieving exciting results, it remains a special case of structure-from-motion methods, which inherently require a certain level of camera baseline and are unable to handle videos with excessively low view coverage.
%
Besides, accurately discovering the correct placement of the neural bones and skinning weights requires a video to demonstrate the movable parts with real-world motions, thus {\ourmethod} cannot hallucinate bones and articulations in the completely invisible regions.
We acknowledge these limitations and aim to address them in future work.



{
    \small
    \bibliographystyle{ieeenat_fullname}
    \bibliography{main}
}

\end{document}